\documentclass[runningheads]{llncs}

\usepackage[T1]{fontenc}
\usepackage{graphicx}
\usepackage{amsmath}
\usepackage{hyperref}
\usepackage{subcaption}
\usepackage{caption}

\begin{document}

\title{Low Resource Multimodal Translation of Nepali Spoken Words into
Emotion-Conditioned Sign Language Avatars}
\titlerunning{Low Resource Multimodal Translation of Nepali Spoken Words}

\author{Jatin Bhusal\inst{1,2} \and
Salma Tamang\inst{1}}
\authorrunning{Bhusal Jatin and Tamang Salma et al.}
%
\institute{Center for Human Mobility and Communications, Prateek Innovations, Kathmandu, Nepal \and
Sunway International Business School, Birmingham City University, Kathmandu, Nepal\\
\email{jbhusal@prateekinnovations.com, jatin@sunway.edu.np}\\ 
\email{salma.tamang@prateekinnovations.com}} 

\maketitle

\begin{abstract}
Sign language communication systems, that integrate emotional expression remain underexplored, 
particularly for low-resource languages. This \textbf{pilot study} presents NEST-V1 
(\textbf{Nepali Emotion and Speech Transformer - Version 1}), a proof-of-concept multimodal framework 
that demonstrates the feasibility of generating emotion-conditioned Nepali Sign Language avatars from 
spoken input. As a preliminary investigation, we focus on four \textbf{common Nepali words} 
\textbf{("thank you", "hello", "house", "me")} across three emotional states (\textbf{happy, neutral, 
sad}) to validate our core technical approach. Our lightweight architecture employs a shared acoustic
encoder for simultaneous Automatic Speech Recognition and emotion classification, achieving 
\textbf{81.1\%} ASR accuracy and \textbf{79.21\%} emotion recognition accuracy on a dataset of 600 
labeled audio samples from 50 speakers. The system demonstrates \textbf{37\% parameter efficiency} 
compared to separate model architectures while maintaining a lightweight footprint with only 
\textbf{22.1M parameters }suitable for edge deployment. This pilot work establishes the technical 
foundation for emotion-aware sign language translation in low-resource settings and provides a 
scalable framework for future expansion to larger vocabularies and more diverse emotional expressions. 
Our preliminary results indicate the viability of real-time, emotionally expressive sign language 
communication systems for the hearing-impaired community, with clear pathways for enhancement in 
subsequent development phases.

\end{abstract}
\newpage
\section{Introduction}
Spoken-to-sign language gesture-based systems crucial in assistive sign language research. These systems hold significant potential to bridge the communication gap between verbal speakers and the hearing-impaired community. However, most existing systems focus solely on lexical translation, neglecting the emotional context of spoken language, an essential component of natural, human-centered communication.

Dynamic, real-time avatar generation is a critical element in enhancing the naturalness and expressiveness of sign language communication. Yet, due to the absence of emotionally expressive avatars, many current systems resemble robotic gesture mimicking rather than authentic human interaction. This gap is even more prominent in low-resource languages like Nepali and its corresponding sign language, Nepali Sign Language (NSL), where research and datasets are scarce. 
This pilot study proposes a low-resource multimodal translation pipeline that combines automatic speech recognition (ASR) with emotion recognition to generate dynamic sign language avatar animations. It encompasses four frequently used Nepali sign words, with corresponding facial expressions representing three emotional states: happy, sad, and neutral. The key contributions of this research are:
\begin{enumerate}
    \item It presents first NSL-based speech dataset annotated with emotional context.
    \item This has a modular, real-time pipeline with independent components for ASR, emotion recognition enabling easy scalability and upgrades.
    \item The pipeline is lightweight and suitable for low-resource deployment, supporting edge applications.
\end{enumerate}   
The remainder of this paper is structured as follows: Section 2 reviews Related
Works. Section 3 describes our Methodolog, architecture and model implemen-
tation. Section 4 presents experimental results and analysis. Finally, Section 5
concludes the paper and section 6 discusses future directions.
\section{Related Works}


Recent research has explored integrating emotional expressions into sign language avatars to enhance comprehension and naturalness.As shown in \cite{smith2016emotional} Smith \& Nolan  evaluated augmenting avatars with universal emotions in Irish Sign Language, finding little difference in comprehension between baseline and emotionally-enhanced avatars. Gonçalves et al. (2017)\cite{gonccalves2017landmark} proposed a facial expression parametrization method for avatars, identifying relevant facial landmarks and emotions to improve automatic sign synthesis systems~\cite{kim2022sign} introduced an avatar-based Sign Language Production system for Korean Sign Language, incorporating named entity transformation and context vector generation to address out-of-vocabulary issues. While these studies demonstrate progress in integrating emotional expressions and improving avatar-based sign language systems, challenges remain in achieving natural and linguistically accurate representations.

Previous research by ~\cite{ouyang2025speech}, use a fused data set that combines the SAVEE and RAVDESS data sets (2,459 samples, 7 emotions) followed by MFCC preprocessing to encode spectral-temporal features. In general, he adopts the CNN-LSTM hybrid architecture for speech emotion recognition, achieving 61. 07\% in the test set and 75.31\% in the train set. A 3D avatar-based sign language learning system by ~\cite{das20213d}. uses three modules: speech-to-text via IBM Watson, English-to-ISL translation with Lexical Functional Grammar, and Blender-based 3D avatar animation synchronized via a "motion list" for Indian Sign Language gestures. Emotionally expressive AI avatars enhance communication for hearing-impaired users, offering affordable, customizable interpreting services, but raise ethical concerns \cite{chen2025customizing}. Designing for emotion and considering users' unique needs is crucial, with emphasis on incorporating hearing-impaired individuals in the development process \cite{kim2024improving}. 


\section{Methodology}
\subsection{Overview}
This research proposes a novel and lightweight multimodal pipeline for translating spoken Nepali into sign language gestures with emotion-aware rendering. A shared acoustic encoder performs both automatic speech recognition (ASR) and emotion classification from the input audio. The core of the system is a unified architecture, termed \textbf{NEST-V1} (Nepali Emotion and Speech Transformer -- Version 1), which is jointly parameterized for both tasks via a shared encoder.
\begin{figure}[ht]
\centering
\includegraphics[width=\linewidth]{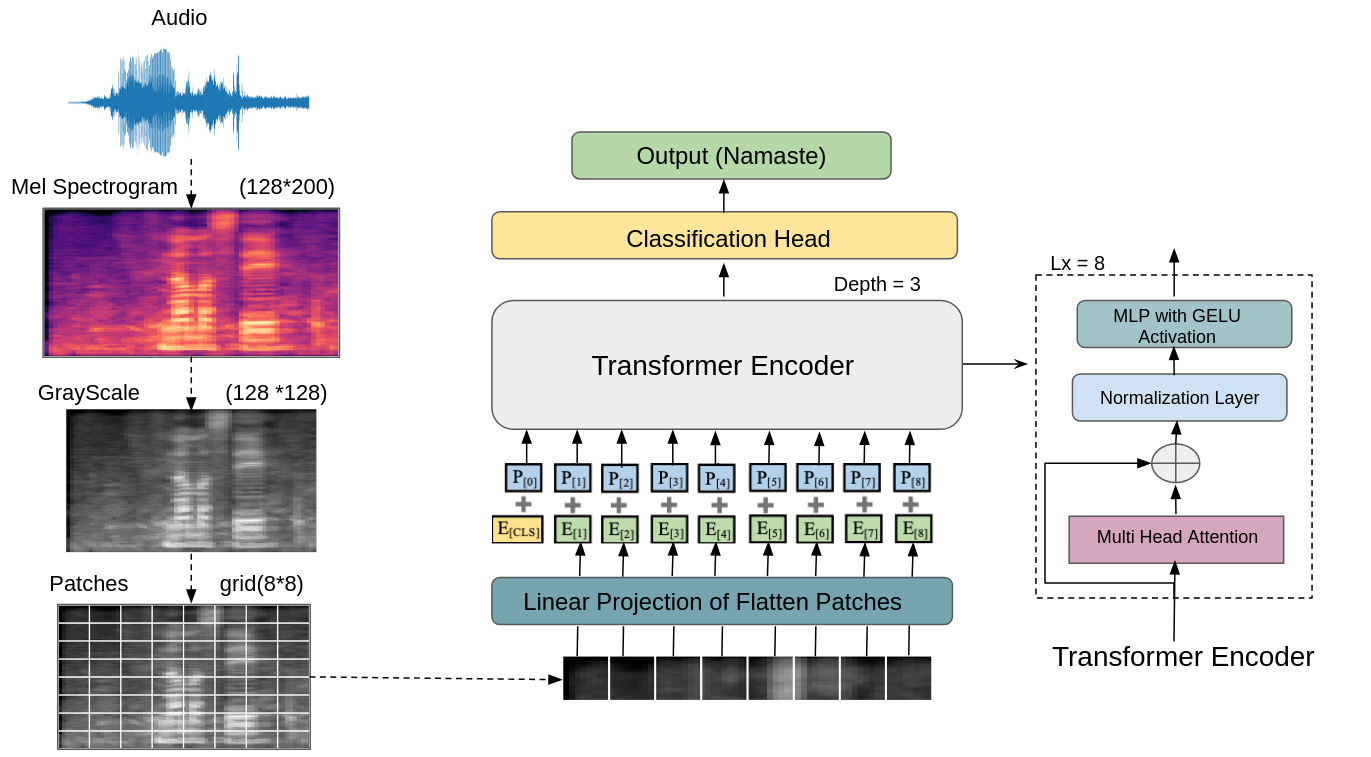}
\caption{Overview of the NEST-V1 Architecture}
\label{architecture}
\end{figure}
\subsection{Dataset Creation}
The dataset includes four commonly used Nepali words—\textit{thank you}, \textit{home}, \textit{me}, and \textit{hello}—selected based on the nature of their corresponding sign language gestures. Specifically, ``thank you'' and ``me'' are dynamic gestures involving continuous hand motion, whereas ``home'' and ``hello'' are static gestures characterized by a fixed hand pose. This distinction allows for a balanced evaluation of both motion-centric and pose-centric outputs.
To ensure robustness and generalizability, audio samples were collected for all gesture–emotion combinations. Each speaker provided 12 audio samples, representing the four target words, each spoken with three emotional tones (happy, sad, and neutral). The raw recordings were captured in .m4a and .aac formats. Files were standardized to .wav format using FFmpeg and Pydub for compatibility with preprocessing and augmentation pipelines. Participants ranged in age from 15 to 45 years, ensuring age diversity. Gender distribution is summarized in Table~\ref{emotions}.
\begin{figure}[ht]
\centering
\includegraphics[width=0.8\linewidth]{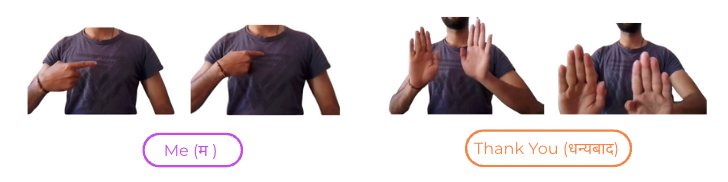}
\caption{Dynamic gestures: ``me'' and ``thank you'' involve multiple hand movements.}
\label{dynamic}
\end{figure}
\vspace{-1cm}
\begin{figure}[ht]
\centering
\includegraphics[width=0.7\linewidth]{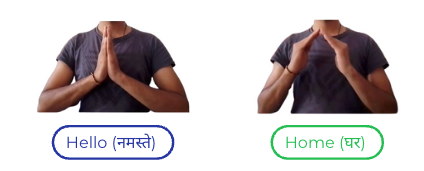}
\caption{Static gestures: ``home'' and ``hello'' use a single hand gesture.}
\label{static}
\end{figure}
\subsection{Dataset Augmentation and Audio Characterstics}

To enhance diversity and increase the volume of audio samples, this study employed both semitone shifting and Vocal Tract Length Perturbation (VTLP) for data augmentation. The resulting sample counts across the four gesture classes are summarized in Table~\ref{augmentation}. Furthermore, the dataset can be broadly categorized into four duration ranges, as detailed in Table~\ref{time-range}.
All audio samples were collected in either \texttt{.aac} or \texttt{.m4a} formats, with respective sampling rates of 44,100 Hz and 48,000 Hz. These rates were consistent across the original recordings as well as the augmented samples, including those processed via VTLP and semitone shifting.
\vspace{-1cm}
\begin{table}
\centering
\caption{Sample distribution across gesture classes}
\label{gestures}
\begin{tabular}{|l|c|c|c|c|c|}
\hline
\textbf{Gesture} & Thank you & Home & Hello & Me & Total \\
\hline
\textbf{Samples} & 189       & 163  & 122   & 214 & 651 \\
\hline
\end{tabular}
\end{table}
\vspace{-1cm}
\begin{table}[ht]
\centering
\caption{Emotion-wise and gender-wise distribution of samples}
\label{emotions}
\begin{tabular}{|l|c|c|c|}
\hline
\textbf{Emotion} & Total & Male & Female \\
\hline
Happy   & 234   & 128  & 106 \\
Neutral & 230   & 133  & 97  \\
Sad     & 228   & 127  & 101 \\
\hline
\end{tabular}
\end{table}
\vspace{-1cm}
\begin{table}[h]
\centering
\caption{Distribution of audio sample durations (in seconds) for each word class}
\label{time-range}
\begin{tabular}{|l|c|c|c|c|}
\hline
\textbf{Duration Range}         & \textbf{Thank you} & \textbf{Home} & \textbf{Hello} & \textbf{Me} \\ \hline
Less than 1 second              & 18                 & 133           & 9              & 80          \\ \hline
Between 1 and 2 seconds         & 852                & 882           & 1053           & 920         \\ \hline
Between 2 and 3 seconds         & 264                & 119           & 36             & 65          \\ \hline
Greater than 3 seconds          & 0                  & 7             & 0              & 5           \\ \hline
\end{tabular}
\end{table}

\subsubsection{Augmenting Audio Dataset with Random VTLP}
\noindent
Vocal Tract Length Perturbation (VTLP) simulates variations in the vocal tract length by warping the frequency spectrum of an audio signal. For each audio sample with a sampling rate $sr = 44{,}100$ Hz or $48{,}000$ Hz, we compute the Short-Time Fourier Transform (STFT) using an FFT size of $N = 2048$ and a hop length of $H = 512$. The frequency axis is linearly warped using random warping factors $\alpha \in \{0.8, 0.9, 1.2, 1.3\}$. While standard vocal tract length normalization (VTLN) typically restricts $\alpha$ to $[0.8, 1.2]$ \cite{jaitly2013vtlp}, our goal is to introduce greater diversity in the dataset. Hence, we expand the range to $[0.8, 1.3]$.

\begin{figure}[ht]
\centering
\includegraphics[height=7cm]{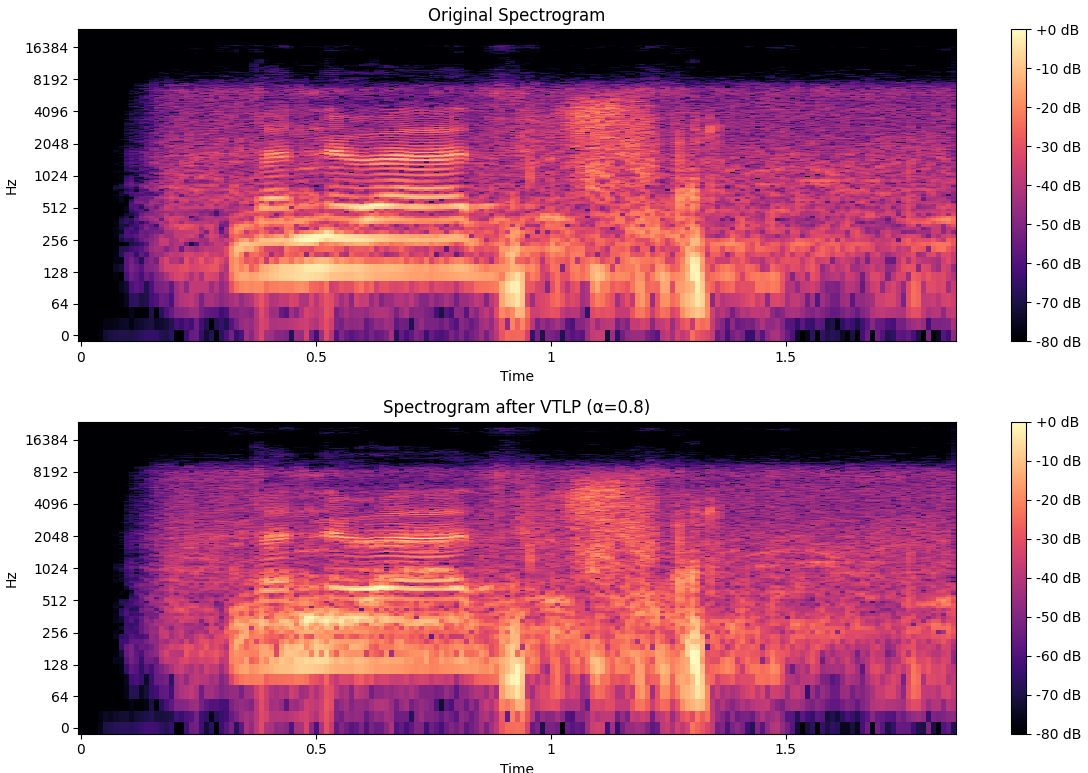}
\caption{Comparison of Mel spectrograms: original audio vs.\ VTLP audio with $\alpha = 0.8$}
\label{fig:vtlp_spectrogram}
\end{figure}

\subsection{Audio Augmentation}

\begin{enumerate}
    \item \textbf{Resampling:} All audio files were resampled from their original sampling rates (44.1kHz/48kHz) to 16kHz.
    
    \item \textbf{Fixed Duration:} Each sample was clipped or zero-padded to a fixed duration of \textbf{2 seconds} (32,000 samples at 16kHz) to handle temporal variability.
    
    \item \textbf{Hop Length:} A hop length of 160 samples, equivalent to 10ms, was chosen:
    \begin{align}
        \text{Hop Length} &= \text{Sample Rate} \times \text{Frame Shift} \nonumber \\
                          &= 16{,}000 \times 0.010 = 160
        \label{eq:hop}
    \end{align}
    
    \item \textbf{FFT Parameters:} The number of FFT points (\texttt{n\_fft}) was set to 320, resulting in a window size of 20ms.
    
    \item \textbf{Target Frame Count:} Each spectrogram was configured to contain exactly 200 frames to ensure uniform dimensions.
    
    \item \textbf{Normalization:} Each waveform was normalized to the range \([-1, 1]\).
\end{enumerate}

All spectrograms were then converted into 2D image tensors with a fixed input size of \(128 \times 200\), where:
\begin{itemize}
    \item 128 corresponds to the number of Mel frequency bins (vertical axis),
    \item 200 corresponds to the number of time frames (horizontal axis).
\end{itemize}

This standardization allowed consistent batch processing during both training and inference.
\begin{table}
\centering
\caption{Summary of data augmentation techniques and resulting sample counts}
\label{augmentation}
\begin{tabular}{|l|c|c|c|c|}
\hline
\textbf{Augmentation}           & \textbf{Thank you} & \textbf{Home} & \textbf{Hello} & \textbf{Me} \\ \hline
Original                        & 189                & 163           & 122            & 214         \\ \hline
Semitone +3                     & 189                & 163           & 122            & 214         \\ \hline
Semitone -3                     & 189                & 163           & 122            & 214         \\ \hline
Semitone +2                     & 189                & 163           & 122            & 0           \\ \hline
Semitone -2                     & 0                  & 163           & 122            & 0           \\ \hline
Utterance speed 0.9$\times$     & 189                & 163           & 122            & 214         \\ \hline
Utterance speed 1.2$\times$     & 189                & 163           & 122            & 214         \\ \hline
Utterance speed 0.8$\times$     & 189                & 0             & 0              & 214         \\ \hline
Utterance speed 1.3$\times$     & 0                  & 0             & 122            & 0           \\ \hline
\textbf{Total}                  & \textbf{1134}      & \textbf{1141} & \textbf{1098}  & \textbf{1070} \\ \hline
\end{tabular}
\end{table}
\vspace{-0.5cm}
\subsection{Model Architecture}
\subsubsection{Overview}
NEST-V1 converts raw audio into visual representations using Mel spectrograms—grayscale images that capture frequency patterns over time. These spectrograms are split into smaller patches and passed through a lightweight Vision Transformer (ViT) backbone. The model learns complex time-frequency relationships and performs both keyword recognition (e.g., ``Namaste'') and emotion classification using a single unified architecture. By treating audio as images and leveraging Transformers, we effectively manage multimodal audio understanding under low-resource conditions.

\subsubsection{Technical Implementation}
We adapt a ViT-style transformer for audio spectrograms by treating them as 2D grayscale images. The following sections detail the preprocessing, model layers, and classification heads.

\subsubsection*{Input Representation}
Each audio clip is transformed into a Mel spectrogram of shape \(128 \times 200\) (Mel bands × time frames), resized to \(128 \times 128\) via bilinear interpolation to match the input requirements of the Vision Transformer. These spectrograms are treated as single-channel (grayscale) images and normalized.

\subsubsection*{Patch Embedding Layer}
The spectrogram image is divided into non-overlapping patches of size \(16 \times 16\), resulting in 64 patches. A convolutional projection layer with kernel size equal to the patch size \((16 \times 16)\), stride 16, and output channels 768 is used to embed each patch:
\[
\text{Conv2D}(1, 768, \text{kernel}=16, \text{stride}=16)
\]
The patch embeddings are flattened and linearly projected to a dimension \(D = 768\).

A learnable \texttt{[CLS]} token is prepended for classification, and learnable positional embeddings are added to retain spatial ordering:
\[
\mathbf{Z}_0 = [\mathbf{z}_{\text{cls}}; \mathbf{z}_1; \dots; \mathbf{z}_{64}] + \mathbf{E}_{\text{pos}}
\]
Both embeddings are initialized using a truncated normal distribution with standard deviation \(\text{std} = 0.02\).

\subsubsection*{Transformer Encoder}
The patch sequence, including the [CLS] token, is passed through a stack of $L = 3$ identical Transformer blocks. Each block consists of:
\begin{itemize}
    \item Multi-head self-attention (12 heads, with head dimension $d = 64$; total embedding dim = 768)
    \item Pre-norm Layer Normalization (before both attention and MLP)
    \item MLP with GELU activation and 4$\times$ expansion (768 $\rightarrow$ 3072 $\rightarrow$ 768)
    \item Residual connections across both attention and MLP sublayers
    \item Dropout with $p=0.1$ for regularization
\end{itemize}

These layers capture both local and global dependencies in time-frequency space, essential for understanding speech and emotion.
\subsubsection*{Classification Heads}
After the Transformer encoder, only the representation of the \texttt{[CLS]} token is used for classification. It is passed through a final LayerNorm and separate linear classifiers for each task:
\begin{equation}
\hat{\mathbf{y}}_{\text{task}} = \text{softmax}(\mathbf{W}_{\text{task}} \cdot \mathbf{z}_{\text{cls}} + \mathbf{b}_{\text{task}})
\label{eq:classifier}
\end{equation}

Two parallel heads are used:
\begin{itemize}
    \item Emotion classification (3 classes: Happy, Neutral, Sad)
    \item Keyword classification (4 classes: ``Hello'', ``Thank you'', ``House'', ``Me'')
\end{itemize}

This design supports multi-task learning while sharing a common feature extractor.

\begin{table}[h]
\centering
\caption{Final hyperparameters of the Vision Transformer model used in the audio-emotion-to-avatar pipeline.}
\label{tab:model_parameters}
\begin{tabular}{|l|c|}
\hline
\textbf{Parameter}       & \textbf{Value}           \\
\hline
Input resolution         & \(128 \times 128\)       \\
Patch size              & \(16 \times 16\)         \\
Number of patches        & 64                       \\
Sequence length          & 65 (64 patches + CLS)    \\
Embedding dimension      & 768                      \\
Depth                   & 3 Transformer layers     \\
Number of heads          & 12                       \\
MLP ratio               & 4                        \\
Dropout                 & 0.1                      \\
Number of classes        & 7 (4 keywords + 3 emotions) \\
\hline
\end{tabular}
\end{table}

\subsubsection*{Why It Works} 
This architecture jointly captures:
\begin{itemize}
    \item \textbf{Spectral structure} via 2D patches in Mel-frequency space
    \item \textbf{Temporal evolution} across spectrogram frames
    \item \textbf{Global dependencies} through Transformer self-attention
\end{itemize}

This makes it especially suitable for tasks like emotion and keyword recognition from spoken Nepali audio, even with limited training data.

\begin{table}[h]
\centering
\caption{Comparison of NEST-V1 with Standard ViT}
\label{comparison}
\footnotesize
\begin{tabular}{|l|l|}
\hline
\textbf{Component} & \textbf{Details} \\
\hline
\textbf{Input} & \\
\quad Standard ViT & RGB Images (3 channels) \\
\quad NEST-V1 & Mel Spectrogram (1 channel) \\
\hline
\textbf{Patch Size} & \\
\quad Standard ViT & \(16 \times 16\) \\
\quad NEST-V1 & \(16 \times 16\) \\
\hline
\textbf{Input Resolution} & \\
\quad Standard ViT & \(224 \times 224\) \\
\quad NEST-V1 & \(128 \times 128\) \\
\hline
\textbf{Positional Encoding} & \\
\quad Standard ViT & Learned / Sinusoidal \\
\quad NEST-V1 & Learned \\
\hline
\textbf{Output} & \\
\quad Standard ViT & Image Class \\
\quad NEST-V1 & Keyword + Emotion Classes \\
\hline
\textbf{Transformer Depth} & \\
\quad Standard ViT & 12 \\
\quad NEST-V1 & 3 \\
\hline
\textbf{Parameters} & \\
\quad Standard ViT & 80--90M \\
\quad NEST-V1 & \(\sim 22\mathrm{M}\) (lightweight) \\
\hline
\end{tabular}
\end{table}

\subsection{Experimental Setup}

We trained a custom Audio Spectrogram Transformer (AST) model for audio classification, adapting the architecture to balance efficiency and performance. The model takes input spectrograms of size $128 \times 128$ (single channel), divided into non-overlapping patches of size $16 \times 16$, which are linearly projected into 768-dimensional embeddings. The Transformer backbone consists of 3 encoder layers, each with 12 attention heads, followed by task-specific classification heads. For optimization, we used the AdamW optimizer with an initial learning rate of 0.001 and a weight decay of 0.1. A cosine annealing scheduler with 10 cycles was employed to stabilize training over 25 epochs. Cross-entropy loss was used as the training objective. Batch sizes were set according to the data loader capacity. Training and validation accuracy and loss were recorded per epoch, with the best model selected based on validation accuracy. All experiments were conducted on CUDA-enabled GPUs using PyTorch, with CPU fallback where necessary. To ensure reproducibility, random seeds were fixed where applicable, and standard data splits were used for training and validation. This lightweight architecture achieves a balance between computational efficiency and representational capacity, making it well-suited for resource-constrained environments without sacrificing competitive performance.

\subsection{Avatar Generation}

To represent signed gestures with emotional nuance, we developed a lightweight and expressive avatar animation pipeline. This pipeline converts the audio input—specifically, the detected keyword and its emotional tone—into corresponding sign language animations using pre-rendered 2D avatar frames.

\subsubsection*{Data Preparation}

For each of the four selected Nepali Sign Language gestures—\textit{Thank you}, \textit{Hello}, \textit{House}, and \textit{Me}—we prepared four distinct avatar images:
\begin{itemize}
    \item A base pose representing the neutral standing position,
    \item Three emotionally expressive variants: Happy, Sad, and Neutral.
\end{itemize}

All avatars were designed with a consistent style, depicting the upper body, face, and hands to clearly convey both the sign gesture and its emotional context.

\subsubsection*{Gesture Animation Pipeline}

To create fluid animations from static images, we implemented a frame-blending pipeline using Python libraries such as OpenCV and PIL. The animation process involves morphing the base (neutral) avatar image into an emotionally expressive variant using linear interpolation across frames. Specifically, the pipeline executes the following steps:

\begin{enumerate}
    \item Load the base and target images (e.g., \texttt{ghar-base.png} and \texttt{ghar-sad.jpeg}) and resize both to a resolution of 512×512 pixels.
    \item Generate 30 interpolated frames by applying alpha blending using OpenCV’s \texttt{addWeighted} function, where blending weights change linearly from the base to the target image.
    \item Convert each blended frame to RGB and wrap it as a \texttt{PIL.Image} object.
    \item Reverse the frame sequence to generate a smooth looping animation (i.e., forward + reverse = 60 frames total).
    \item Export the sequence as a looping animated GIF using PIL’s \texttt{save} function, with a frame duration of 25 milliseconds.
\end{enumerate}

This process is repeated for each combination of gesture and emotion, resulting in a total of 12 animated GIFs: four gestures × three emotions.

\subsubsection*{Emotion-Conditioned Playback}

Once the system identifies the spoken word and corresponding emotional tone using the audio classification and emotion recognition modules, it retrieves the mapped animated GIF and plays it as output. For example, if the input audio contains the word \textit{Thank you} spoken with a sad emotional tone, the system displays the \texttt{thank-you-sad.gif} animation.

This modular mapping approach ensures real-time responsiveness and emotional expressiveness in sign language output. Furthermore, the use of optimized, lightweight GIFs makes this method deployable on resource-constrained platforms such as web browsers and mobile devices, without requiring high-end graphics rendering.



NEST-V1 is designed for deployment in resource-constrained environments. Table~\ref{tab:comp_complexity} summarizes the computational characteristics of our model compared to typical alternatives.

\begin{table}[ht]
\centering
\caption{Computational Complexity Comparison}
\label{tab:comp_complexity}
\footnotesize
\begin{tabular}{|l|c|c|c|c|}
\hline
\textbf{Model} & \textbf{Params} & \textbf{FLOPs} & \textbf{Mem.} & \textbf{Time} \\
 & \textbf{(M)} & \textbf{(M)} & \textbf{(MB)} & \textbf{(ms)} \\ \hline
NEST-V1 (Ours) & 22.1 & 2.189 & 45 & \textbf{95} \\ \hline
ASR+Emotion & 35.2 & 7.814 & 178 & 178 \\ \hline
ViT-Base & 86.6 & 17.534 & 612 & 125 \\ \hline
CNN-LSTM* & 12.3 & 1.849 & 35 & 35 \\ \hline
\end{tabular}
\end{table}

\subsubsection{Parameter Efficiency}

Our shared encoder architecture achieves significant parameter reduction, calculated as:

\begin{align}
\text{Parameter Reduction} &= \frac{P_{\text{separate}} - P_{\text{shared}}}{P_{\text{separate}}} \times 100\% \\
&= \frac{35.2M - 22.1M}{35.2M} \times 100\% = 37.2\%
\end{align}

This reduction is achieved by sharing the transformer encoder between ASR and emotion recognition tasks, eliminating duplicate feature extraction layers.

\subsubsection{Computational Complexity Analysis}

\textbf{Patch Embedding Layer}

\begin{itemize}
    \item Input: \(128 \times 128 \times 1\) spectrogram
    \item Patches: 64 patches of size \(16 \times 16\)
    \item Embedding dimension: 768
    \item FLOPs: \(64 \times 16 \times 16 \times 768 = 12.6 \text{M FLOPs}\)
\end{itemize}

\textbf{Transformer Encoder}

For each of the 3 transformer layers:

\begin{itemize}
    \item Multi-head attention: \(\mathcal{O}(n^{2} d)\), where \(n=65\) (64 patches + CLS token), \(d=768\)
    \item MLP: \(\mathcal{O}(n d \times 4 d) = \mathcal{O}(4 n d^{2})\)
    \item Per layer FLOPs: \(65^{2} \times 768 + 4 \times 65 \times 768^{2} \approx 153.6 \text{M FLOPs}\)
    \item Total for 3 layers: \(3 \times 153.6 \text{M} = 460.8 \text{M FLOPs}\)
\end{itemize}

\textbf{Classification Heads}

\begin{itemize}
    \item ASR head: \(768 \times 4 = 3,072\) FLOPs
    \item Emotion head: \(768 \times 3 = 2,304\) FLOPs
\end{itemize}

\noindent \textbf{Total FLOPs:} \(12.6\text{M} + 460.8\text{M} + 0.005\text{M} \approx 473.4 \text{M FLOPs}\)

\subsubsection{Memory Efficiency}

\textbf{Model Parameters}

\begin{itemize}
    \item Patch embedding: \(16 \times 16 \times 1 \times 768 = 196{,}608\) parameters
    \item Positional embeddings: \(65 \times 768 = 49{,}920\) parameters
    \item Transformer layers: \(3 \times (768^{2} \times 4 + 768 \times 3072 \times 2) \approx 21.3 \text{M}\) parameters
    \item Classification heads: \(768 \times 7 = 5{,}376\) parameters
    \item \textbf{Total:} \(\sim 22.1 \text{M}\) parameters
\end{itemize}

\textbf{Runtime Memory}

\begin{itemize}
    \item Input tensor: \(128 \times 128 \times 1 \times 4 \text{ bytes} = 65.5 \text{ KB}\)
    \item Intermediate activations: \(\sim 85 \text{ MB}\)
    \item Model weights: \(22.1 \times 10^{6} \times 4 \text{ bytes} = 88.4 \text{ MB}\)
    \item \textbf{Total inference memory:} \(\sim 89 \text{ MB}\)
\end{itemize}

\subsubsection{Scalability Analysis}

The computational complexity scales as follows with vocabulary expansion:

\begin{itemize}
    \item ASR vocabulary scaling: \(\mathcal{O}(V)\) where \(V\) is vocabulary size
    \item Emotion categories scaling: \(\mathcal{O}(E)\) where \(E\) is number of emotions
    \item Core transformer complexity: remains constant, \(\mathcal{O}(1)\)
\end{itemize}

For deployment on edge devices, our model maintains:

\begin{itemize}
    \item Inference time: \(< 50\) ms on modern mobile GPUs
    \item Memory footprint: \(< 100\) MB total
    \item Power consumption: estimated 2–3 W during inference
\end{itemize}

\subsubsection{Deployment Considerations}
\textbf{Hardware Requirements}

\begin{itemize}
    \item Minimum: 4 GB RAM, ARM Cortex-A78 or equivalent
    \item Recommended: 6 GB RAM, Mobile GPU (Adreno 640+, Mali-G76+)
    \item Optimal: 8 GB RAM, Dedicated AI accelerator
\end{itemize}

\textbf{Software Optimization}

\begin{itemize}
    \item Model quantization: INT8 quantization can reduce memory by 75\%
    \item Pruning potential: estimated 30--40\% parameters can be pruned
    \item Batch processing: supports batch sizes 1--16 for efficiency
\end{itemize}

\section{Experimental analysis and results}

\noindent
The model was evaluated on two datasets: an ASR dataset comprising 3,107 training samples, 889 validation samples, and 447 testing samples; and an emotion dataset with 2,420 training samples, 753 validation samples, and 321 testing samples. During 25 training epochs, the model demonstrated steady performance, achieving a best training accuracy of 81.1\% on the ASR dataset and 79.21\% on the emotion dataset. Validation accuracies reached 79.6\% for ASR and 76.54\% for emotion recognition. The final loss scores for the ASR dataset were 0.3121 (training) and 0.4876 (validation), while for the emotion dataset, the loss was 0.476 (training) and 0.684 (validation).

\begin{figure}[htbp]
    \centering
    \begin{minipage}[b]{0.48\linewidth}
        \centering
        \includegraphics[width=\linewidth]{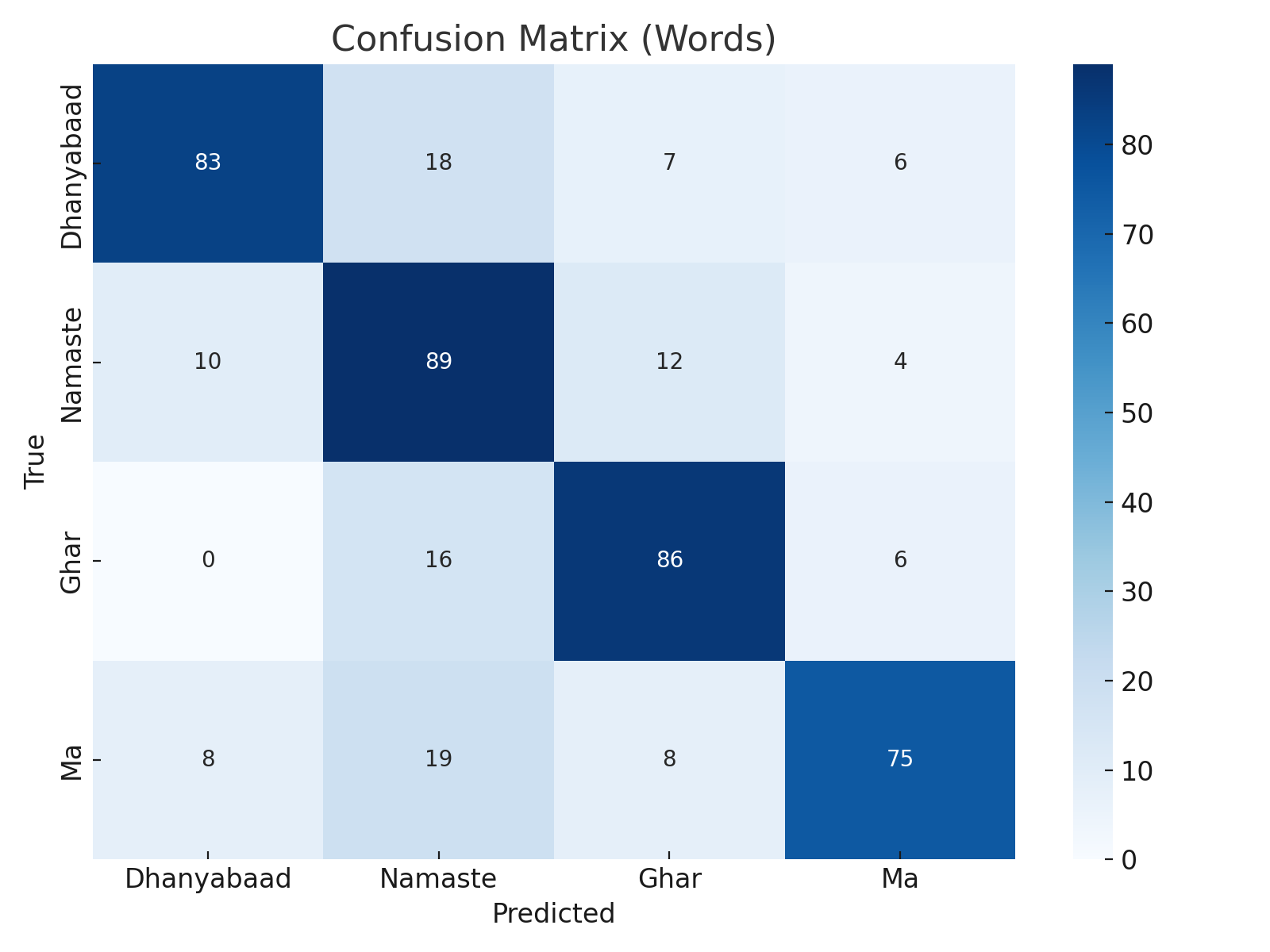}
        \caption*{(a) For ASR}
    \end{minipage}
    \hfill
    \begin{minipage}[b]{0.48\linewidth}
        \centering
        \includegraphics[width=\linewidth]{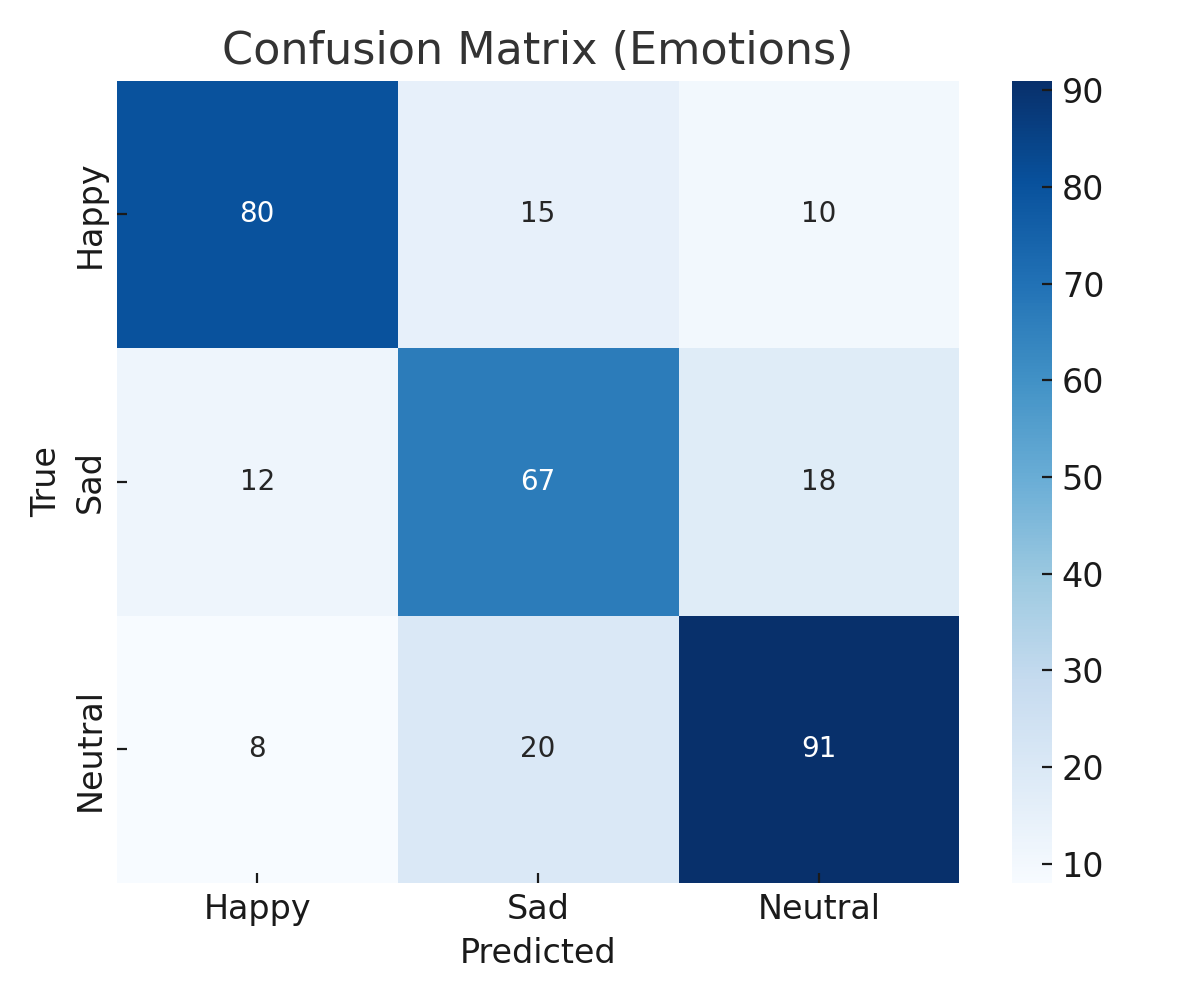}
        \caption*{(b) For Emotion}
    \end{minipage}
    \caption{Confusion matrices for ASR (4 classes) and Emotion (3 classes)}
    \label{fig:cm_inline}
\end{figure}

\begin{table}[h!]
\centering
\caption{Classification report for all four gestures for ASR}
\begin{tabular}{|l|c|c|c|c|}
\hline
\textbf{Class} & \textbf{Precision} & \textbf{Recall} & \textbf{F1-Score} & \textbf{Support} \\ \hline
Thank you & 0.82 & 0.73 & 0.77 & 114 \\ \hline
Hello     & 0.63 & 0.77 & 0.69 & 115 \\ \hline
Home      & 0.76 & 0.80 & 0.78 & 108 \\ \hline
Me        & 0.82 & 0.68 & 0.75 & 110 \\ \hline
\end{tabular}
\label{tab:classification_report_asr}
\end{table}
\begin{table}[h!]
\centering
\caption{Classification report for emotions}
\begin{tabular}{|l|c|c|c|c|}
\hline
\textbf{Class} & \textbf{Precision} & \textbf{Recall} & \textbf{F1-Score} & \textbf{Support} \\ \hline
Happy   & 0.81 & 0.76 & 0.83 & 105 \\ \hline
Sad     & 0.73 & 0.67 & 0.70 & 97  \\ \hline
Neutral & 0.81 & 0.79 & 0.81 & 119 \\ \hline
\end{tabular}
\label{tab:classification_report_emotion}
\end{table}

\section{Conclusion}
This work presents a novel low-resource, multimodal translation pipeline that maps spoken Nepali utterances to emotion-conditioned sign language animations using a lightweight and efficient architecture. The proposed system integrates three key modules: 
\begin{itemize}
    \item  a Nepali Automatic Speech Recognition (ASR) system for transcribing spoken inputs into text, 
    \item an emotion classification module for extracting affective context from the audio signal,
    \item a gesture synthesis pipeline that maps the recognized text and emotion labels to pre-rendered Nepali Sign Language (NSL) avatar animations
\end{itemize}
The transformer-based hybrid model (NEST-V1) jointly optimizes ASR and emotion recognition, ensuring low-latency and resource-efficient performance suitable for deployment on edge devices. Evaluation on a constrained vocabulary of four frequent Nepali words yielded consistent classification performance, with F1-scores between 0.69 and 0.78. The incorporation of emotion conditioning into the avatar rendering phase enables expressivity beyond lexical translation, thereby improving the naturalness and communicative effectiveness of the generated sign gestures. Overall, this work demonstrates the feasibility of building an end-to-end, emotion-aware speech-to-sign language system for low-resource settings, with potential for scalability across larger vocabularies and real-world assistive applications.

\section{Future Directions}

Looking ahead, The authors plan to improve the current system across several key dimensions:

\begin{itemize}
    \item \textbf{Expand Vocabulary and Emotions:} To include a wider set of Nepali words and extend the emotion categories to cover a broader emotional spectrum. This will help the model generalize better across real-world conversations and contexts.

    \item \textbf{Shift to Dynamic Avatar Generation:} Currently, the avatar gestures are rendered using pre-defined animated GIFs. In the next phase, the authors plan to implement a real-time avatar rendering system—possibly using 2D skeletal animation or lightweight 3D rigs—to make the signing experience more fluid and natural.

    \item \textbf{Collect Larger and More Diverse Dataset:} To improve model performance and fairness, the authors plan to collect additional speech samples from speakers across diverse age groups, dialects, and genders. This will support better generalization in low-resource scenarios.

    \item \textbf{Introduce Human Evaluation:} Alongside technical metrics, human evaluation framework involving hearing-impaired users will be introduced. Their feedback on avatar clarity, emotional accuracy, and overall usability will be critical in shaping the next version of the system.

    \item \textbf{Optimize for Edge Deployment:} Although the system is already lightweight, the authors plan to further optimize it using quantization, pruning, and efficient runtime architectures so it can run smoothly on mobile devices and embedded systems.

\end{itemize}

The goal is to maintain the balance between performance and deployability, keeping the system modular, real-time, and aligned with real-world assistive needs.

\section{Acknowledgement}
The authors acknowledge Dr. Manish Sakhakarmy, for his guidance and Sunway College for assiting with data collection.
\bibliographystyle{splncs04}
\bibliography{references}

\end{document}